\documentclass[11pt]{article}

\usepackage[final]{acl}

\usepackage{times}
\usepackage{latexsym}
\usepackage{amsmath,amssymb,amsfonts}
\usepackage{bbold}
\usepackage{makecell}
\usepackage{multirow}
\usepackage{colortbl}
\usepackage{xcolor}
\usepackage{pdfpages}
\usepackage{hyperref}
\usepackage{booktabs}
\usepackage{pifont}
\usepackage{subcaption}
\usepackage{enumitem}
\allowdisplaybreaks[4]

\definecolor{personapromptcolor}{HTML}{EDB8B0}
\definecolor{auditorpromptcolor}{HTML}{E3B87F}
\definecolor{repairpromptcolor}{HTML}{B2D3A4}
\definecolor{tablecolor}{HTML}{AABCDB}
\definecolor{faithcolor}{HTML}{EAB67A}
\definecolor{ablationcolor}{HTML}{DACFE5}
\definecolor{caseblue}{HTML}{0070C0}
\definecolor{casered}{HTML}{C00000}
\definecolor{casegreen}{HTML}{00B050}
\definecolor{casestudyboxgrey}{HTML}{BFC1C0}

\usepackage[most]{tcolorbox}
\tcbuselibrary{skins, breakable}
\newtcolorbox{personapromptbox}[1]{
    breakable,
    colback=white,
    colframe=personapromptcolor,
    arc=4pt,
    boxrule=0.8pt,
    title=#1,
    coltitle=white,
    colbacktitle=personapromptcolor,
    fonttitle=\bfseries
}
\newtcolorbox{auditorpromptbox}[1]{
    breakable,
    colback=white,
    colframe=auditorpromptcolor,
    arc=4pt,
    boxrule=0.8pt,
    title=#1,
    coltitle=white,
    colbacktitle=auditorpromptcolor,
    fonttitle=\bfseries
}
\newtcolorbox{repairpromptbox}[1]{
    breakable,
    colback=white,
    colframe=repairpromptcolor,
    arc=4pt,
    boxrule=0.8pt,
    title=#1,
    coltitle=white,
    colbacktitle=repairpromptcolor,
    fonttitle=\bfseries
}
\newtcolorbox{casestudybox}[1]{
    enhanced,
    breakable,
    colback=white,
    colframe=casestudyboxgrey,
    arc=4pt,
    boxrule=0.8pt,
    title=#1,
    coltitle=white,
    colbacktitle=casestudyboxgrey,
    fonttitle=\bfseries
}

\usepackage[T1]{fontenc}

\usepackage[utf8]{inputenc}

\usepackage{microtype}

\usepackage{inconsolata}

\usepackage{graphicx}

\title{Verify Before You Commit: Towards Faithful Reasoning in LLM Agents via Self-Auditing}

\author{
 \textbf{Wenhao Yuan\textsuperscript{1}},
 \textbf{Chenchen Lin\textsuperscript{2}},
 \textbf{Jian Chen\textsuperscript{1}},
 \textbf{Jinfeng Xu\textsuperscript{1}},
\\
 \textbf{Xuehe Wang\textsuperscript{2}},
 \textbf{Edith Cheuk Han Ngai\textsuperscript{1,\thanks{Corresponding author.}}}
\\
 \textsuperscript{1}The University of Hong Kong,
 \textsuperscript{2}Sun Yat-sen University
\\
   \href{mailto:wenhao.yuan@connect.hku.hk}{wenhao.yuan@connect.hku.hk}, \href{mailto:chngai@eee.hku.hk}{chngai@eee.hku.hk}
}

\begin{document}
\maketitle
\begin{abstract}
In large language model (LLM) agents, reasoning trajectories are treated as reliable internal beliefs for guiding actions and updating memory. However, coherent reasoning can still violate logical or evidential constraints, allowing unsupported beliefs repeatedly stored and propagated across decision steps, leading to systematic behavioral drift in long-horizon agentic systems. Most existing strategies rely on the consensus mechanism, conflating agreement with faithfulness. In this paper, inspired by the vulnerability of unfaithful intermediate reasoning trajectories, we propose \textbf{S}elf-\textbf{A}udited \textbf{Ve}rified \textbf{R}easoning (\textsc{SAVeR}), a novel framework that enforces verification over internal belief states within the agent before action commitment, achieving faithful reasoning. Concretely, we structurally generate persona-based diverse candidate beliefs for selection under a faithfulness-relevant structure space. To achieve reasoning faithfulness, we perform adversarial auditing to localize violations and repair through constraint-guided minimal interventions under verifiable acceptance criteria. Extensive experiments on six benchmark datasets demonstrate that our approach consistently improves reasoning faithfulness while preserving competitive end-task performance.
\end{abstract}

\section{Introduction}

\begin{figure}[t]
\centerline{\includegraphics[width=0.48\textwidth, trim=0 0 0 0,clip]{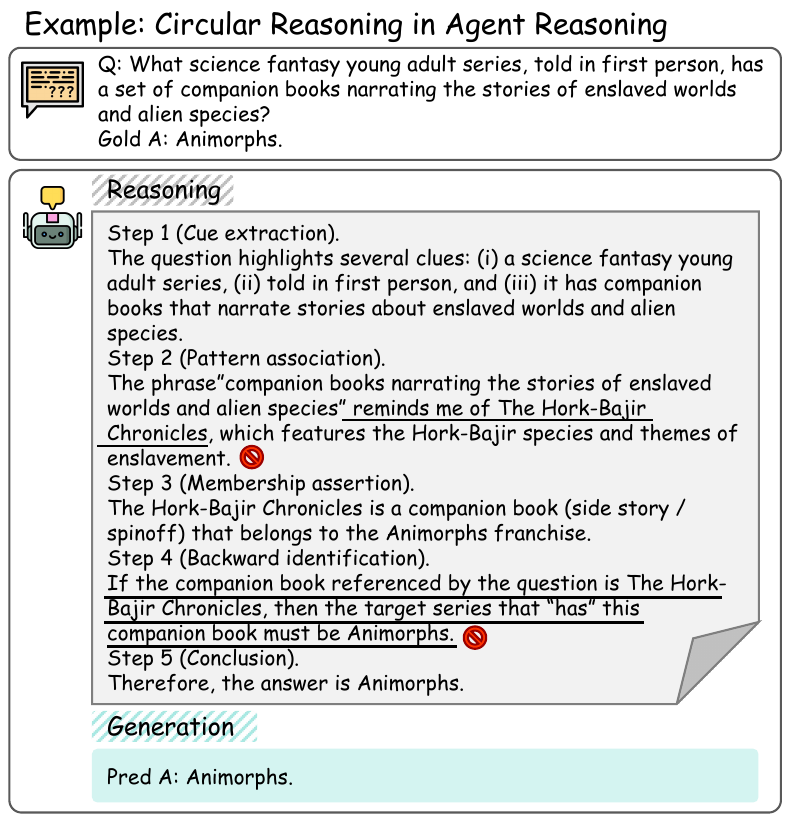}}
\caption{Demonstration of unfaithful agent reasoning. The agent outputs the correct answer `\texttt{Animorphs}', but its multi-step reasoning process is logically invalid, as an unverified intermediate assumption ``\texttt{The\! phrase} $\ldots$ \texttt{reminds\! me\! of\! The\! Hork-Bajir\! Chronicles}" is used to derive the conclusion that it already presupposes. This failure mode differs fundamentally from unfaithful CoT, where the reasoning is merely an explanatory artifact, while unfaithful reasoning in the agent determines the following behavior and final decision.}
\label{intro_fig}
\end{figure}

Large Language Models (LLMs) are increasingly deployed as autonomous \textit{agents} that plan, reason, and act over extended horizons. Beyond generating answers, LLM agents maintain internal reasoning trajectories for guiding tool invocation, action commitment, and memory updates across decision steps. With the widespread adoption of reasoning paradigms, such as Chain-of-Thought (CoT)~\citep{wei2022chain}, trajectories are generally regarded as interpretable representations of the agent's internal state. However, coherent reasoning traces are fragile for decision-making~\citep{lam2025codecrash}. Agents may generate seemingly fluent and structured reasoning, yet violate logical or evidential constraints, reflecting a lack of faithful reasoning~\citep{zhao2025chain, xu2025softcot}. Such violations are difficult to diagnose from final-task success alone, since correct outcomes can arise from chance, redundancy, or downstream correction, masking the underlying reasoning failure~\citep{chang2025survey, kim2024qube}, as shown in Figure~\ref{intro_fig}. Unlike single-turn Question Answering (QA), where reasoning can be post hoc and disposable, the agent's reasoning outputs are repeatedly used, amplified, and written into memory~\citep{an2025thread, jiang2025qa, tang2025chemagent}. Consequently, unfaithful belief states (e.g., unsupported inferences or hidden assumptions) can propagate, bias decisions, and trigger costly actions in closed-loop agent systems~\citep{chakraborty2025hallucination}. The risk is not merely incorrect answers, but systematic behavioral drift driven by unfaithful internal beliefs.

In agentic systems, existing methods have been adopted to manage uncertainty before committing internal reasoning states to actions, such as self-consistency~\citep{wan2025reasoning, xie2024calibrating}, multi-agent debate~\citep{liang2026multi, liang2024encouraging}, which maintain multiple candidate reasoning trajectories and rely on consensus-based aggregation for acted belief determination~\citep{zhang2025debate4math}. Nevertheless, they rest on the problematic premise that consensus is faithfulness. In practice, multiple sampled trajectories frequently share the same implicit assumptions or inference templates, resulting in structurally correlated yet unfaithful belief states that are repeatedly selected, further reinforced by majority voting, and committed to memory~\citep{ke2025survey}. Additionally, most existing methods interact with reasoning at the level of surface text rewriting~\citep{shu2024rewritelm}, without identifying the logical constraints that the specific reasoning step violates, and verifiable acceptance criteria for committing corrected belief states. These limitations reveal that current LLM agents lack an objective of ensuring reasoning faithfulness before action commitment, raising a key question: \textit{How can an LLM agent verify the reasoning faithfulness without relying on final-task accuracy or consensus?} 

To address these challenges, we propose \textbf{S}elf-\textbf{A}udited \textbf{Ve}rified \textbf{R}easoning (\textsc{SAVeR}), a novel framework for enhancing reasoning faithfulness in LLM agents. Rather than relying on final-task outcomes, \textsc{SAVeR} explicitly models the faithfulness of intermediate reasoning steps. To mitigate correlated failure patterns in belief generation, the agent generates a persona-conditioned coalition to elicit structurally diverse candidate belief states and reduce repeated unfaithful templates. It then selects beliefs in a faithfulness-relevant structure space via a quality-aware diversity kernel and $k$-DPP sampling, followed by adversarial auditing that localizes violations into auditable diagnostics. Finally, \textsc{SAVeR} introduces a constraint-guided minimal counterfactual repair protocol that edits only localized failure slices under verifiable acceptance criteria, iterating audit–repair until the belief passes all checks before being committed to actions or memory. Our key contributions are as follows:
\begin{itemize}
\item We reveal the overlooked issue of reasoning faithfulness in LLM agents and identify the challenges of verifying intermediate beliefs before action commitment.
    
\item We introduce \textsc{SAVeR}, a novel self-auditing framework that verifies and repairs intermediate reasoning trajectories in agents.
    
\item We conduct extensive experiments on multiple public datasets to demonstrate the effectiveness of our approach in improving reasoning faithfulness.
\end{itemize}

\section{Related Work}

\subsection{Faithful Reasoning in LLMs}
LLMs have shown strong performance on reasoning tasks. Existing work has explored the prompting strategies, most notably CoT prompting~\citep{wei2022chain, kojima2022large}. Extensions such as program-based reasoning~\citep{chen2023program, zhang2024natural, jiang2024urbanllm, wang2024chain} and least-to-most prompting~\citep{zhou2023least, arora2023adapt} further structure these steps and improve task accuracy~\citep{ma2025should}. However, improved reasoning performance does not necessarily imply faithful reasoning: empirical studies show that generated chains of thought may rely on spurious signals, and intervening on them has limited impact on model predictions, suggesting that such explanations are post-hoc rationalizations~\citep{lyu2023faithful, turpin2023language}.

\begin{figure*}[t]
\centerline{\includegraphics[width=1.0\textwidth, trim=0 5 0 0,clip]{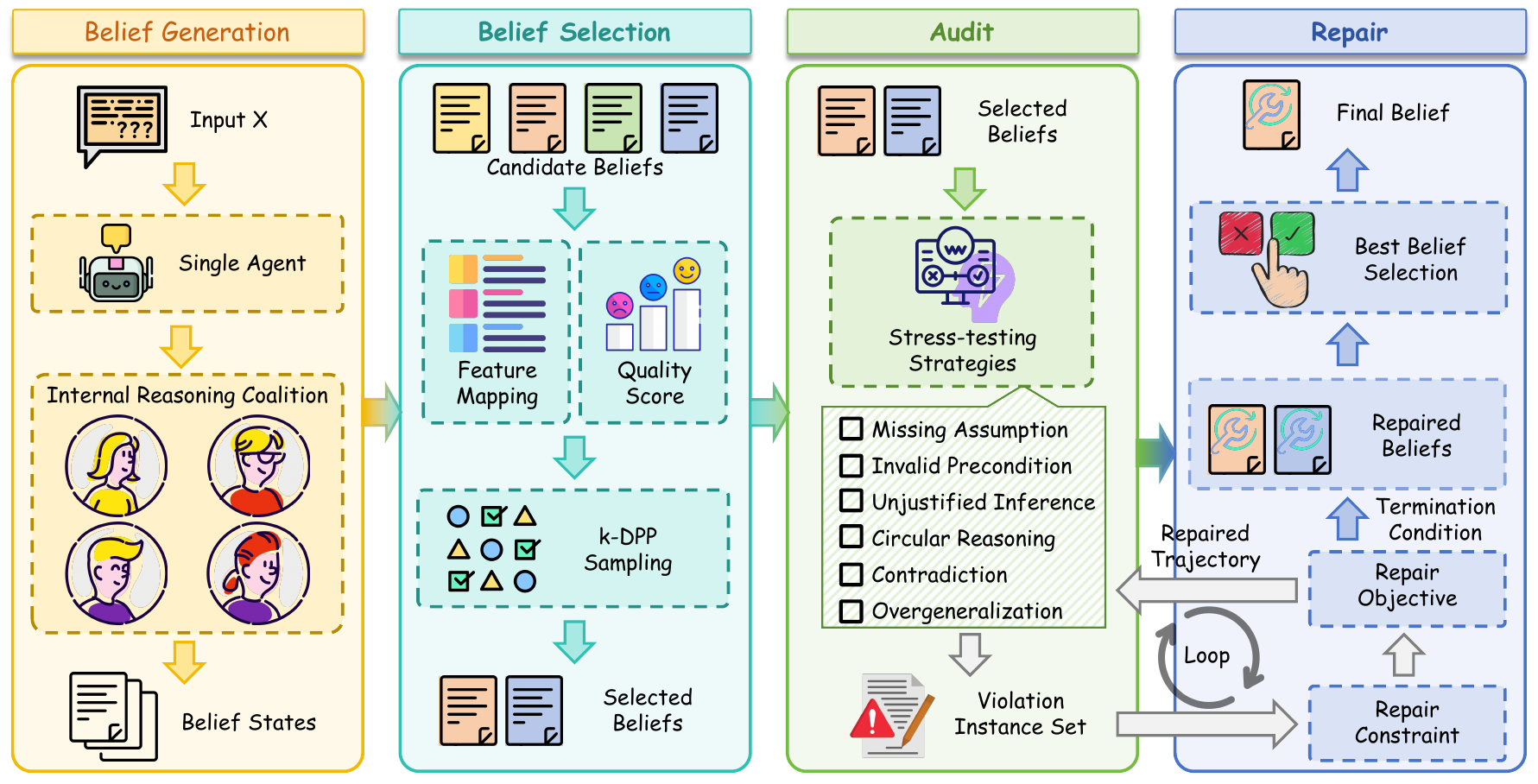}}
\caption{An overview of our proposed \textsc{SAVeR} framework. This diagram illustrates the overall closed-loop workflow, highlighting the end-to-end process from belief generation and selection to auditing and iterative repair. }
\label{framework}
\end{figure*}

Motivated by these concerns, a growing body of work has focused on evaluating reasoning faithfulness in LLMs~\citep{luo2024reasoning, paul2024making, luo2025graph}. Existing approaches include counterfactual interventions on reasoning traces~\citep{paul2024making, ding2024rationale, joshi2024cold}, causal probing methods~\citep{chi2024unveiling, roy2025causal}, and targeted diagnostics for chain-of-thought faithfulness~\citep{li2025towards}. Beyond evaluation, several methods improve faithful reasoning through verification~\citep{dougrez2025assessing} or self-consistency~\citep{wan2025reasoning}. 

Despite these advances, most studies on faithful reasoning focus on LLMs. In agentic settings, however, reasoning trajectories function as persistent belief states to guide downstream decisions, allowing erroneous beliefs to accumulate, propagate, and trigger costly actions over long horizons. LLM-centric methods to faithful reasoning are insufficient for agentic decision-making, underscoring the need for agent-specific mechanisms.

\subsection{Faithfulness-Aware Reasoning in Agents}
An emerging research perspective frames LLMs as agents that perform multi-step reasoning through planning, tool use, and interaction with external environments~\citep{hong2025data, chen2025locagent, xu2025amem}. These frameworks expose explicit reasoning trajectories to support complex decision-making in interactive settings~\citep{yao2023react, yang2024swe}. However, explicit reasoning does not guarantee faithfulness: empirical studies demonstrate that agent-generated plans or tool-use rationales may not causally determine outcomes, but instead serve as plausible post-hoc justifications~\citep{barez2025chain}. 

Despite these observations, existing approaches for mitigating unfaithful agent reasoning remain post-hoc or outcome-driven. Many methods improve robustness by sampling multiple reasoning trajectories or applying external critics, without explicitly verifying whether intermediate reasoning steps are supported by the agent’s available evidence at the time of decision-making~\citep{liang2024abseval, kostka2025towards}. Moreover, numerous approaches operate at surface-level trajectory rewriting or consensus aggregation, which is insufficient to identify structurally correlated yet unsupported belief states that can repeatedly pass heuristic checks~\citep{fu2025agentrefine,grotschla2025agentsnet}. Consequently, current agentic frameworks lack a mechanism for auditing and verifying internal belief states before action, allowing unfaithful reasoning to be propagated and stored in memory.

\section{Methodology}
In this section, we introduce the \textsc{SAVeR} framework for faithful reasoning in agentic systems. The complete workflow is shown in Figure~\ref{framework}.  We first formalize reasoning faithfulness in \S~\ref{model_faithfulness}, and then describe persona-conditioned belief generation in \S~\ref{belief_generation} and structure-aware belief selection in \S~\ref{belief_selection}. We present our reasoning audit mechanism in \S~\ref{reasoning_audit} and introduce the repair procedure that iteratively corrects unfaithful beliefs in \S~\ref{output_repair}.

\subsection{Modeling Faithful Reasoning in Agent} \label{model_faithfulness}
We consider an agent that generates a multi-step internal reasoning trajectory and then commits to external actions. While improving interpretability, it also introduces the challenge of \textit{reasoning faithfulness}: intermediate reasoning steps may not be fully supported by the information available during decision-making, motivating an explicit formulation of reasoning faithfulness.

Given an input task $x$, we model the agent's internal reasoning process as a sequence of discrete steps $\tau = (s_{1}, \ldots, s_{L})$, where $L$ denotes the length of the reasoning trajectory and each step $s_{l}$ represents a local claim, inference, assumption, or intermediate conclusion produced by the agent. To quantify whether a reasoning step is justified under the information available to the agent, we introduce the \textit{support function} $\Gamma(s_{l} \mid x, \mathcal{H}_{l}, \mathcal{E}_{l}) \in [0,1]$, where the reasoning history $\mathcal{H}_{l} = (s_{1},\dots,s_{l-1})$ and $\mathcal{E}_{l}$ represents the accessible evidence at step $l$, including retrieved documents, tool outputs, or environment observations. Thus, we define the \textit{trajectory-level unfaithfulness rate} as
\begin{align}
U(\tau) = \frac{1}{L} \sum_{l=1}^{L} \mathbb{I} [\Gamma(s_{l} \mid x, \mathcal{H}_{l}, \mathcal{E}_{l}) < \epsilon],
\end{align}
where $\mathbb{I}[\cdot]$ denotes the indicator function and a reasoning step is considered \textit{unfaithful} if its support score falls below a predefined threshold $\epsilon$.


\subsection{Persona-Conditioned Belief Generation} \label{belief_generation}
Unfaithful reasoning in agents typically arises from repeatable and structurally correlated failure patterns, making naively sampled reasoning traces unreliable. Rather than improving final answer accuracy by generating more traces, our goal is to promote \textit{structural diversity} among candidate belief states, so that distinct reasoning modes and failure triggers are explicitly exposed. 

For the given input $x$, we instantiate an \textit{internal reasoning coalition} consisting of $M$ reasoning personas by $\mathcal{A}=\{a_{1}, a_{2}, \dots, a_{M}\}$, which models a \textit{single} LLM agent’s internal cognitive diversity, where each persona corresponds to a distinct \textit{structural reasoning bias}, e.g., assumption-first vs.\ evidence-first. Each persona $a_i$ is instantiated via persona-specific instruction constraints and reasoning templates. Let $\mathcal{Y}=\mathcal{C}\times\mathcal{R}$ denote the belief space, where $\mathcal{C}$ is the claim space and $\mathcal{R}$ is the space of reasoning trajectories. Persona $a_i$ then produces belief $y_i = G(x;a_i) \in \mathcal{Y}$. We denote $y_i = (c_i, r_{i})$, where the persona's final claim or decision $c_i \in \mathcal{C}$ and $r_{i} =\{s_{i,1},\dots,s_{i,L_i}\} \in \mathcal{R}$ denotes the full reasoning trajectory with $L_i$ steps and is treated as a \textit{candidate belief state}. 

\subsection{Structure-Aware Belief Selection} \label{belief_selection}
To select diverse subsets of belief states, we define a structural feature mapping $\phi: r_{i} \mapsto \phi(r_{i})\in\mathbb{R}^{d}$, designed as a proxy for reasoning faithfulness-relevant structure. We decompose it as
\begin{align}
\phi(r_{i})= [g(r_{i}), p(r_{i}), v(r_{i}), s(r_{i}) ]^\top,
\end{align}
where \textit{granularity features} $g(r_{i})$ quantify step granularity and potential skipping risk; \textit{assumptive features} $p(r_{i})$ reflect how assumptions are introduced and managed, capturing missing/implicit premises and whether assumptions are properly scoped; \textit{verification features} $v(r_{i})$ measure verification behaviors; \textit{structural-type features} $s(r_{i})$ describe the global organization of reasoning. 

We introduce a lightweight quality scoring function $q(y_i;x)$, which provides coarse filtering of candidate belief states and removes traces that violate minimal usability constraints (e.g., nonsensical steps or internally inconsistent conclusions). Then, we define a \textit{quality-aware diversity kernel} matrix $I\in\mathbb{R}^{M\times M}$ with entries
\begin{align} \label{kernel_matrix}
I_{ij} = \exp(\beta\,\tilde q_i) \exp(\beta \tilde q_j) \kappa(\phi(r_{i}), \phi(r_{j})),
\end{align}
where $\beta$ controls the quality weighting strength, $\tilde{q}_i$ denotes a normalized $q(y_i;x)$, and $\kappa(\cdot,\cdot)$ is a structural similarity kernel applied to $\phi(r_i)$. Given candidate reasoning outputs generated by $M$ personas, the agent selects $K$ belief states for auditing. We adopt a $k$-Determinantal Point Process ($k$-DPP) to sample a subset $S$ of size $K$:
\begin{align} \label{kdpp} 
\mathbb{P}(S)\propto \det(I_S), S\subseteq \{1,\dots,M\},
\end{align}
where $I_S$ denotes the principal submatrix of $I$ defined in Eq.~(\ref{kernel_matrix}) indexed by $S$. The determinant $\det(I_S)$ favors subsets with structurally complementary belief representations in the induced feature space, thereby encouraging the selection of beliefs that exhibit distinct reasoning patterns. As a result, the sampled set avoids allocating auditing capacity to multiple beliefs that violate reasoning faithfulness in similar ways, and instead increases coverage over diverse \textit{unfaithful reasoning} modes.

\subsection{Adversarial Reasoning Audit} \label{reasoning_audit}
Diversity alone does not guarantee reasoning faithfulness, as beliefs may still contain logically unsupported or unjustified steps. To prevent unfaithful beliefs from being committed to actions, we introduce an \textit{adversarial reasoning auditing} procedure to examine belief states, identify faithfulness-violating reasoning steps, and produce structured diagnostic signals for subsequent repair. Notably, \textit{the auditor interrogates the belief state rather than generating or aggregating alternative answers.}

We perform reasoning auditing by applying a collection of complementary stress-testing strategies to each selected belief $y_i$, $i \in S$. Given the reasoning trajectory $r_i$ and input $x$, the auditor operates under an observable context that aggregates stated assumptions, verified intermediate facts, and admissible evidence (e.g., retrieved documents or tool outputs). Each stress-testing strategy audits $r_i$ from distinct logical perspectives and produces structured audit evidence following a fixed schema to ensure auditability and comparability across beliefs (detailed in Appendix~A.2). According to the auditing outcomes, we categorize the faithfulness violations into a type set: $\mathcal{T} = \{\texttt{Missing\_Assumption}, \texttt{Invalid\_Precondition}, \\ \texttt{Unjustified\_Inference}, \texttt{Circular\_Reasoning}, \\ \texttt{Contradiction}, \texttt{Overgeneralization}\}$, which captures the common failure modes in agentic settings, thereby enabling downstream corrective actions (detailed in Appendix~A.1).

For each belief trajectory $r_i$, the auditor outputs a violation instance set $\mathcal{V}(r_i)=\{(t_{i,j}, l_{i,j})\}_{j=1}^{m_i}$, where $m_i = |\mathcal{V}(r_i)|$ denotes the number of violation instances detected in trajectory $r_i$, $t_{i,j}\in\mathcal{T}$ denotes the faithfulness violation type, and $l_{i,j}\in\{1,\dots, L_i\}$ indexes the reasoning step at which the violation is triggered, distinguishing between globally unfaithful beliefs and those that fail only at specific steps. The auditing process operationalizes this notion by identifying steps $s_{i,l}$ for which $\Gamma(s_{i,l} \mid x, \mathcal{H}_{i,l}, \mathcal{E}_{i,l}) < \epsilon$ and mapping each instance to a concrete violation type $t\in\mathcal{T}$. In this way, the violation instance set $\mathcal{V}(r_i)$ can be viewed as a discrete, structured instantiation of support assessment. To represent the belief's faithfulness failure characteristics, we summarize the auditing outcome as an unfaithfulness profile $\mathbf{h}(r_{i}) = [h_{t}(r_{i})]_{t \in \mathcal{T}}$, where $h_{t}(r_{i})$ counts the number of violations of type $t$ in trajectory $r_{i}$.

\begin{table*}[ht]
\centering
\renewcommand\arraystretch{0.9}
\resizebox{1.0\textwidth}{!}{
\begin{tabular}
{c|c|cc|cc|cc|cc|cc|c} 
\toprule[1.2pt]
\multirow{2}{*}{\multirowcell{1.5}{\centering\textbf{Models}}} & \multirow{2}{*}{\multirowcell{1.5}{\centering\textbf{Methods}}} & \multicolumn{2}{c|}{\centering\textbf{HotpotQA}} & \multicolumn{2}{c|}{\centering\textbf{2WikiMHQA}} & \multicolumn{2}{c|}{\centering\textbf{MuSiQue}} & \multicolumn{2}{c|}{\centering\textbf{NQ}} & \multicolumn{2}{c|}{\centering\textbf{Quoref}} & \multicolumn{1}{c}{\centering\textbf{FEVER}}  \\ \cmidrule[0.5pt](l{1pt}r{0pt}){3-13}

& & EM $\uparrow$ & F1 $\uparrow$ & EM $\uparrow$ & F1 $\uparrow$ & EM $\uparrow$ & F1 $\uparrow$ & EM $\uparrow$ & F1 $\uparrow$ & EM $\uparrow$ & F1 $\uparrow$ & EM $\uparrow$ \\ \cmidrule[0.8pt](l{1pt}r{0pt}){1-13}

\multirow{5}{*}{\multirowcell{2}{LLaMA-3.1-8B}}
        & \textsc{Vanilla LM} & 34.8 & 43.6 & 39.6 & 47.3 & 23.8 & 34.6 & 29.5 & 38.2 & 29.4 & 37.7 & 53.7  \\ 

        & \textsc{CoT} & 38.3 & 47.5 & 44.2 & 50.7 & 27.1 & 36.8 & 32.7 & 43.4 & 33.2 & 42.3 & 55.9  \\ 

        & \textsc{MAD} & \underline{43.1} & 51.2 & \textbf{47.9} & \underline{55.4} & 30.9 & \underline{40.8} & \underline{36.6} & \underline{46.9} & 36.3 & 45.2 & 60.7  \\ 

        & \textsc{Self-Refine} & 40.8 & 48.9 & 46.3 & 53.3 & 28.7 & 37.8 & 33.5 & 43.0 & 34.1 & 43.6 & 57.6  \\ 

        & \textsc{B}-2 & 42.9 & \underline{51.3} & 46.7 & 54.4 & \underline{31.0} & 40.6 & 35.9 & 44.9 & \underline{36.7} & \textbf{46.2} & 60.6  \\ 
        
        & \cellcolor{tablecolor}\textsc{SAVeR}(Ours) & \cellcolor{tablecolor}\textbf{43.7} & \cellcolor{tablecolor}\textbf{52.6} & \cellcolor{tablecolor}\underline{47.7} & \cellcolor{tablecolor}\textbf{55.5} & \cellcolor{tablecolor}\textbf{31.8} & \cellcolor{tablecolor}\textbf{42.5} & \cellcolor{tablecolor}\textbf{37.1} & \cellcolor{tablecolor}\textbf{47.8} & \cellcolor{tablecolor}\textbf{37.2} & \cellcolor{tablecolor}\underline{45.7} & \cellcolor{tablecolor}\textbf{61.1}  \\ \midrule[0.8pt]

\multirow{5}{*}{\multirowcell{2}{LLaMA-3.2-3B}}
        & \textsc{Vanilla LM} & 30.6 & 39.1 & 31.6 & 39.4 & 11.7 & 20.8 & 25.1 & 34.8 & 24.4 & 34.6 & 49.3  \\ 

        & \textsc{CoT} & 34.4 & 43.6 & 35.0 & 43.6 & 15.2 & 23.4 & 29.5 & 38.6 & 29.2 & 37.8 & 52.4  \\ 

        & \textsc{MAD} & \underline{37.4} & \underline{45.8} & \underline{39.8} & \underline{47.2} & \underline{18.4} & \underline{28.1} & 33.4 & 42.4 & \textbf{33.5} & \underline{41.2} & \underline{55.6}  \\ 

        & \textsc{Self-Refine} & 34.9 & 44.1 & 36.8 & 44.5 & 17.1 & 26.4 & 31.2 & 39.2 & 29.9 & 39.4 & 53.8  \\ 

        & \textsc{B}-2 & 36.0 & 45.7 & 39.9 & 46.9 & 18.3 & 27.5 & \textbf{34.4} & \underline{42.9} & 32.1 & 40.7 & 54.3  \\ 
        
        & \cellcolor{tablecolor}\textsc{SAVeR}(Ours) & \cellcolor{tablecolor}\textbf{38.3} & \cellcolor{tablecolor}\textbf{47.5} & \cellcolor{tablecolor}\textbf{40.0} & \cellcolor{tablecolor}\textbf{48.6} & \cellcolor{tablecolor}\textbf{18.6} & \cellcolor{tablecolor}\textbf{28.3} & \cellcolor{tablecolor}\underline{33.9} & \cellcolor{tablecolor}\textbf{43.8} & \cellcolor{tablecolor}\underline{33.2} & \cellcolor{tablecolor}\textbf{41.9} & \cellcolor{tablecolor}\textbf{56.4}  \\ \midrule[0.8pt]

\multirow{5}{*}{\multirowcell{2}{Qwen-2.5-7B}}
        & \textsc{Vanilla LM} & 33.9 & 42.3 & 38.4 & 46.0 & 21.9 & 30.5 & 28.2 & 36.9 & 28.4 & 36.5 & 52.7  \\ 

        & \textsc{CoT} & 38.6 & 46.6 & 42.3 & 49.3 & 26.3 & 34.7 & 33.0 & 41.7 & 32.8 & 42.5 & 56.3  \\ 

        & \textsc{MAD} & \underline{42.5} & \underline{50.9} & 46.9 & 54.8 & \textbf{30.8} & \underline{39.3} & \underline{36.2} & 44.6 & \underline{35.3} & \textbf{44.8} & 60.1  \\ 

        & \textsc{Self-Refine} & 39.4 & 48.6 & 44.1 & 52.2 & 27.2 & 36.4 & 34.6 & 43.7 & 33.8 & 42.9 & 58.2  \\ 

        & \textsc{B}-2 & 41.2 & 50.6 & \underline{47.2} & \underline{55.1} & 29.1 & 39.0 & 35.5 & \underline{45.3} & 35.1 & 43.3 & \underline{60.7}  \\ 
        
        & \cellcolor{tablecolor}\textsc{SAVeR}(Ours) & \cellcolor{tablecolor}\textbf{43.1} & \cellcolor{tablecolor}\textbf{51.2} & \cellcolor{tablecolor}\textbf{47.7} & \cellcolor{tablecolor}\textbf{55.8} & \cellcolor{tablecolor}\underline{30.6} & \cellcolor{tablecolor}\textbf{39.4} & \cellcolor{tablecolor}\textbf{36.8} & \cellcolor{tablecolor}\textbf{45.9} & \cellcolor{tablecolor}\textbf{35.6} & \cellcolor{tablecolor}\underline{44.1} & \cellcolor{tablecolor}\textbf{60.9} \\ 

\bottomrule[1.2pt]
\end{tabular}}
\caption{The overall evaluation results of \colorbox{tablecolor}{\textsc{SAVeR}} and other baseline methods on six benchmarks. The best-performed method is marked by \textbf{bold} and the runner-up performing method is marked by \underline{underline}.}
\label{accuracy}
\vspace{-10pt}
\end{table*}

\subsection{Constraint-Guided Belief Repair} \label{output_repair}
Auditing alone does not improve reasoning faithfulness, and full regeneration of new reasoning traces will generally break the causal link between critique and correction, and make it hard to guarantee the removal of originally identified failure. Consequently, we adopt a \textit{minimal counterfactual intervention} principle that only edits the localized failure slices identified, while keeping unaffected steps stable to preserve auditability and prevent unnecessary drift. Specifically, for each violation instance $(t_{i,j}, l_{i,j})\in\mathcal{V}(r_i)$, the auditor returns structured evidence and an acceptance criterion in a fixed schema (detailed in Appendix~A.2), converting subjective critique into faithfulness constraints through explicit acceptance criteria (see Appendix~B). Given the audit output $\mathcal{V}(r_i)$, we define a repair constraint set $\Theta_i=\{\theta_{i,1},\dots,\theta_{i,m_i}\}$, where each constraint $\theta_{i,j}$ encodes a prescribed correction and an explicit criterion for verifying. Let $r_i$ denote the original belief-specific reasoning trajectory to be repaired, and $\tilde{r}_{i}$ the repaired trajectory, computed by solving
\begin{align} \label{repair_objective}
\tilde{r}_{i} = \arg\min_{r} \mathcal{L}_{\mathrm{cons}}(r;\Theta_i) + \lambda \Delta(r,r_{i}),
\end{align}
where the constraint violation loss $\mathcal{L}_{\mathrm{cons}}(r;\Theta_i) = \sum_{j=1}^{m_i} \mathbb{1}[\neg \mathsf{Sat}(r,\theta_{i,j})]$ penalizes failures to satisfy the acceptance criteria implied by $\Theta_i$, $\mathsf{Sat}(r,\theta)$ encodes a concrete and verifiable condition specifying when a violation is resolved, and the minimal edit cost $\Delta(r,r_{i})$ measures the deviation between the repaired and original trajectories, enforcing minimal intervention. Correcting one violation can expose additional latent violations. We thereby iterate auditing and repair until no violation instances remain, i.e., $\mathcal{V}(\tilde r_i)=\emptyset$, the repaired belief is then committed to action and update memory, preventing the propagation of unfaithful reasoning in long-horizon agentic decision-making.

After repairing the audited subset $\{y_i\}_{i\in S}$, the agent selects a belief for execution by maximizing the quality score $q(\cdot;x)$ while penalizing residual unfaithfulness reflected by $\mathbf{h}(\cdot)$:
\begin{equation} \label{final_commit}
i^\star=\arg\max_{i \in S} (q(\tilde{y}_i;x)-\alpha\sum_{t\in\mathcal{T}} w_t h_t(\tilde{r}_i)),
\end{equation}
where $\tilde{y}_i=(\tilde{c}_i,\tilde{r}_i)$ denotes the repaired belief, $w_t$ is a predefined type-dependent severity weight, and $\alpha$ controls the trade-off between superficial quality and verified reasoning faithfulness.

\section{Evaluation}

\subsection{Experimental Setup}

\paragraph{Datasets}
We evaluate our method on six benchmark datasets with various reasoning settings. 
\textbf{Multi-hop QA} integrates information from multiple sources and performs multi-step reasoning. We adopt three benchmarks for this task: HotpotQA~\citep{yang2018hotpotqa}, 2WikiMHQA~\citep{ho2020constructing}, and MuSiQue \citep{trivedi2022musique}. \textbf{Evidence-sensitive QA} focuses on answering questions or verifying claims where correctness critically depends on whether sufficient and appropriate evidence supports the conclusion, prone to unsupported assumptions and unjustified inference steps. We consider Natural Questions (NQ)~\citep{karpukhin2020dense} and FEVER~\citep{thorne2018fever} in this category. \textbf{Local reasoning} tasks resolve referential dependencies within a single context, serving as a baseline for evaluating reasoning faithfulness under minimal structural uncertainty. We choose Quoref~\citep{dasigi2019quoref} for evaluation.

\begin{table*}[t]
\centering
\setlength{\tabcolsep}{3pt}
\resizebox{1.0\textwidth}{!}{
\begin{tabular}{c|cccc|cccc|cccc} 
\toprule[1.2pt]
\multirow{2}{*}{\multirowcell{2}{\centering\textbf{Methods}}} & \multicolumn{4}{c|}{\centering\textbf{HotpotQA}} & \multicolumn{4}{c|}{\centering\textbf{2WikiMHQA}} & \multicolumn{4}{c}{\centering\textbf{MuSiQue}}  \\ \cmidrule[0.5pt](l{1pt}r{0pt}){2-13}

& Avg Viol $\downarrow$ & VFR $\uparrow$ & Post-Res $\downarrow$ & USR $\downarrow$ & Avg Viol $\downarrow$ & VFR $\uparrow$ & Post-Res $\downarrow$ & USR $\downarrow$ & Avg Viol $\downarrow$ & VFR $\uparrow$ & Post-Res $\downarrow$ & USR $\downarrow$ \\ \cmidrule[0.8pt](l{1pt}r{0pt}){1-13}

\textsc{Vanilla LM} & 2.65 & 7.43\% & -- & 46.41\% & 2.83 & 6.58\% & -- & 53.19\% & 3.25 & 5.34\% & -- & 62.63\% \\ 

\textsc{CoT} & 1.98 & 24.89\% & -- & 27.36\% & 2.21 & 17.41\% & -- & 32.11\% & 2.91 & 13.26\% & -- & 37.58\%  \\ 

\textsc{MAD} & 1.33 & 36.74\% & -- & 23.94\% & 1.81 & 32.78\% & -- & 28.82\% & 2.16 & 26.17\% & -- & 36.51\%  \\ 

\cellcolor{faithcolor}\textsc{SAVeR}(Ours) & \cellcolor{faithcolor}0.37 & \cellcolor{faithcolor}81.36\% & \cellcolor{faithcolor}0.05 & \cellcolor{faithcolor}9.12\% & \cellcolor{faithcolor}0.56 & \cellcolor{faithcolor}72.34\% & \cellcolor{faithcolor}0.08 & \cellcolor{faithcolor}13.84\% & \cellcolor{faithcolor}0.83 & \cellcolor{faithcolor}69.38\% & \cellcolor{faithcolor}0.11 & \cellcolor{faithcolor}19.73\%  \\ 

\bottomrule[1.2pt]
\end{tabular}}
\caption{Reasoning Faithfulness Evaluation on Multi-hop QA Benchmarks under LLaMA-3.1-8B.}
\label{faithfulness}
\end{table*}

\paragraph{Baselines}
We compare our method against state-of-the-art baselines. We adopt \textbf{Vanilla LM} \citep{brown2020language} as a direct generation baseline to answer questions without explicit reasoning. To elicit step-by-step reasoning, we include \textbf{CoT}~\citep{wei2022chain}, where the model produces a rationale before answering. We consider deliberation-based inference with \textbf{Multi-Agent Debate (MAD)}~\citep{liang2024encouraging}, which aggregates multiple agents' discussions to form the final answer. For iterative self-improvement, we adopt \textbf{Self-Refine}~\citep{madaan2023self}, where the model alternates between generating and revising based on self-critique. Finally, we include \textbf{Best-of-2 (B-2)}~\citep{papineni2002bleu} to produce two candidate outputs and select the final answer.

\paragraph{Evaluation Metrics}
We utilize two complementary categories of metrics for evaluation. For \textbf{Task-level Performance}, we report Exact Match (EM) and token-level F1, following standard evaluation protocols for QA and verification tasks. For \textbf{Reasoning Faithfulness}, as correct final answers do not necessarily imply faithful reasoning, we additionally evaluate faithfulness at the trajectory level. Based on the audit results, we compute the following faithfulness metrics: (\expandafter{\romannumeral1}) \textit{Average Violations} (Avg Viol),  the mean number of detected faithfulness violations per reasoning trajectory; (\expandafter{\romannumeral2}) \textit{Violation-Free Rate} (VFR), the proportion of trajectories that contain no detected violations; (\expandafter{\romannumeral3}) \textit{Unfaithful Step Rate} (USR), the fraction of reasoning steps within a trajectory that are flagged as unfaithful; (\expandafter{\romannumeral4}) \textit{Post-Repair Residual} (Post-Res) measures the remaining violation rate after the audit–repair procedure is applied. 

\paragraph{Implementation Details}
We conduct our experiments on Qwen 2.5-7B~\citep{bai2025qwen25vltechnicalreport}, LLaMA-3.1-8B, and LLaMA-3.2-3B~\citep{Dubey2024TheL3, touvron2023llama}. All models are used in the zero-shot inference setting, without task-specific fine-tuning. We default person number $M=4$, select $K=2$ candidates for auditing. We define $\beta=1.0$, $\epsilon=0.5$. The audit-repair process is iterated for at most 10 rounds. Faithfulness evaluation is performed with the same auditing protocol for all methods. Violation statistics are computed based on the final reasoning trajectories produced by each method. We run our models on four NVIDIA RTX 4090 GPU devices.

\begin{figure}[t]
\centering
\subfloat{
    \includegraphics[width=0.23\textwidth, trim= 5 5 5 5,clip]{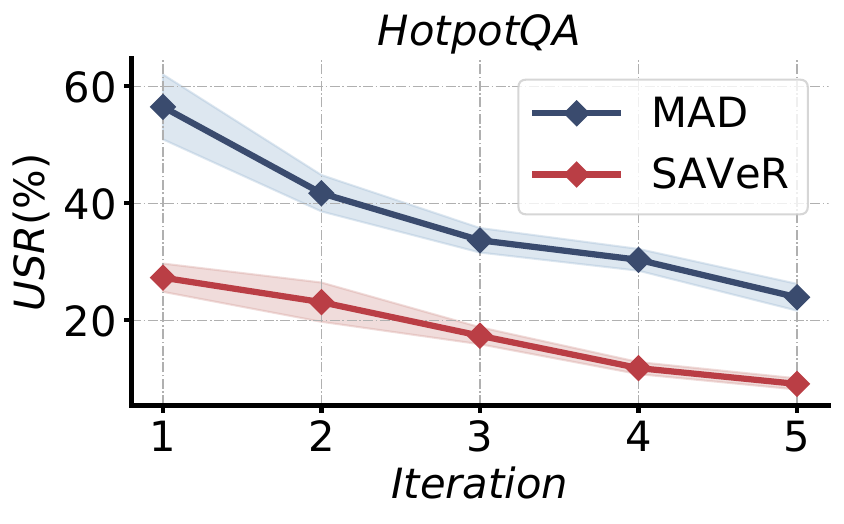}}
\subfloat{
    \includegraphics[width=0.23\textwidth, trim=5 5 5 5,clip]{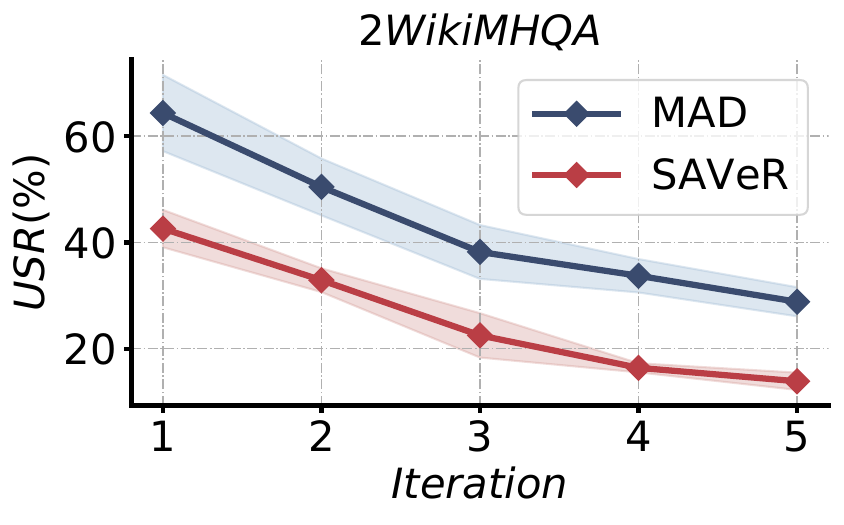}}
\vspace{5pt}
\subfloat{
    \includegraphics[width=0.23\textwidth, trim=5 5 5 5,clip]{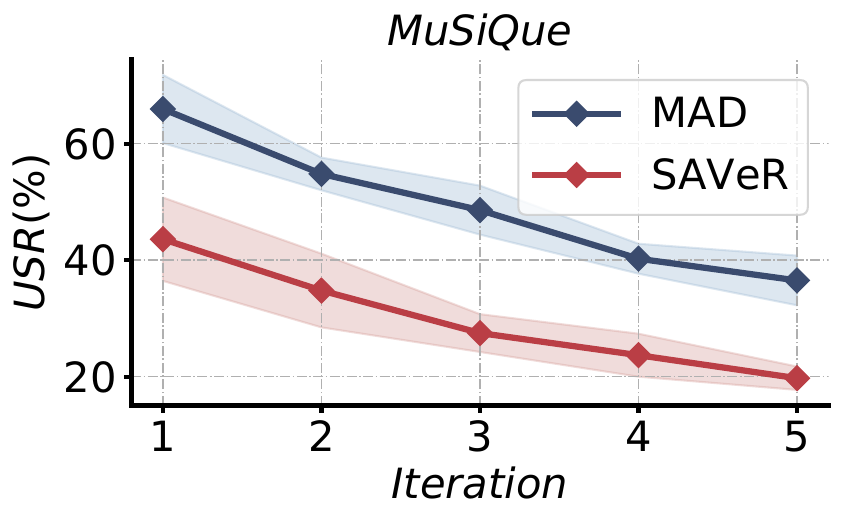}}
\subfloat{
    \includegraphics[width=0.23\textwidth, trim= 5 5 5 5,clip]{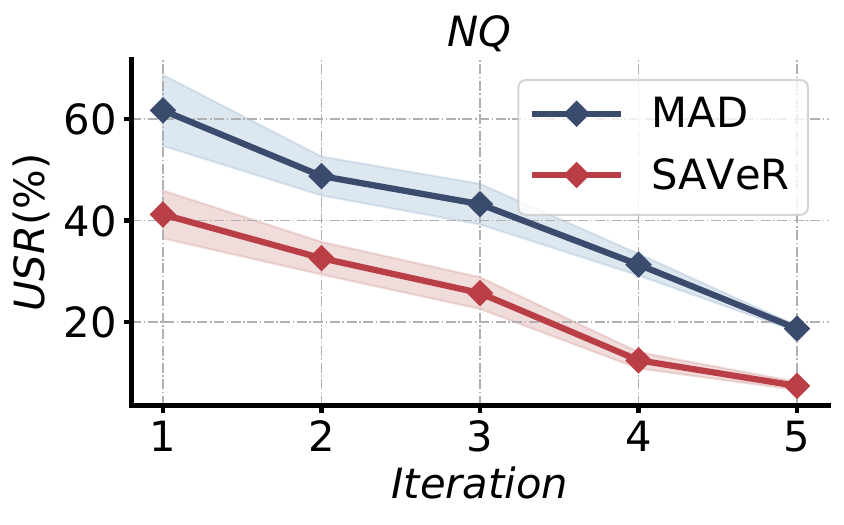}}
\vspace{5pt}
\subfloat{
    \includegraphics[width=0.23\textwidth, trim=5 5 5 5,clip]{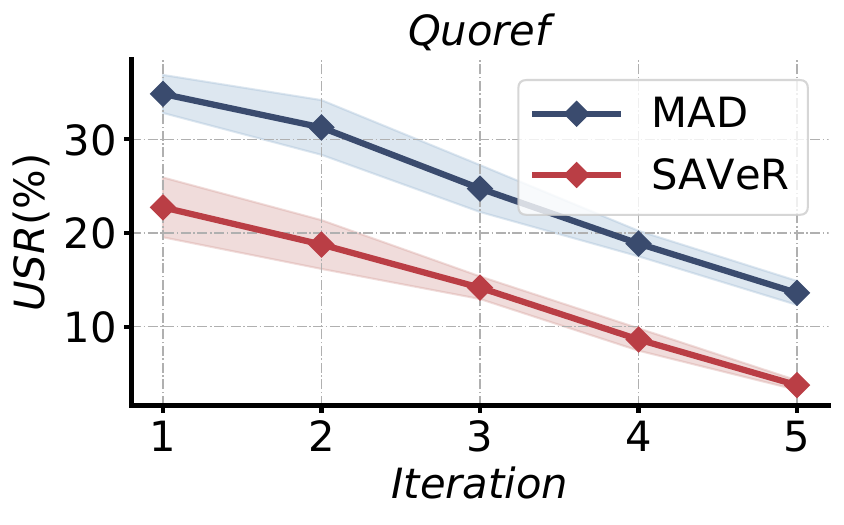}}
\subfloat{
    \includegraphics[width=0.23\textwidth, trim=5 5 5 5,clip]{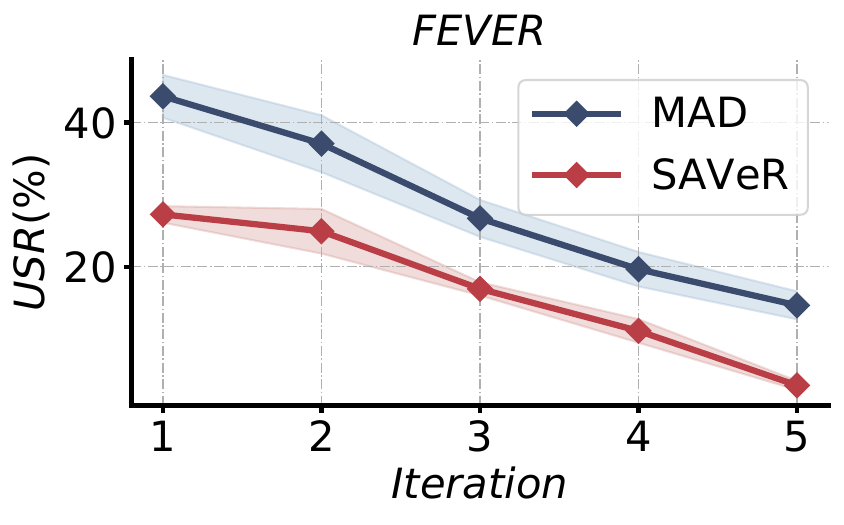}}
\caption{Audit-Repair (A-R) dynamics on HotpotQA. For \textsc{SAVeR}, iterations correspond to one audit-repair cycle. For \textsc{MAD}, iterations denote debates to reduce inconsistencies without verifiable acceptance criteria.}
\label{audit_repair}
\end{figure}

\subsection{Main Results}
Table~\ref{accuracy} reports the overall performance of \textsc{SAVeR} and baselines across six benchmarks under three backbone models, where \textsc{SAVeR} achieves consistently competitive evaluation results. On multi-hop QA benchmarks (HotpotQA, 2WikiMHQA, and MuSiQue), \textsc{SAVeR} demonstrates clear improvements over standard prompting methods and iterative refinement baselines, indicating its effectiveness in handling multi-step reasoning tasks. On evidence-sensitive and single-hop benchmarks (NQ, Quoref, and FEVER), \textsc{SAVeR} also performs competitively, suggesting the superiority of enforcing reasoning verification on performance. We further observe that the performance gains of \textsc{SAVeR} are stable across different model scales. 

We demonstrate the reasoning faithfulness evaluation on three multi-hop QA benchmarks as summarized in Table~\ref{faithfulness}. \textsc{SAVeR} consistently achieves substantially lower Avg Viol and USR, along with markedly higher VFR, compared to all baselines across datasets, indicating a significant reduction in unfaithful intermediate reasoning. In contrast, CoT and MAD alleviate unfaithfulness to a limited extent but still retain a large proportion of violation-prone steps. Moreover, the low Post-Res values of \textsc{SAVeR} indicate that the audit-repair (A-R) process effectively resolves detected violations.

Figure~\ref{audit_repair} illustrates the evolution of USR across iterative refinement for \textsc{SAVeR} and MAD on six benchmarks under LLaMA-3.1-8B. Across all datasets, \textsc{SAVeR} exhibits a faster and more stable reduction in USR, consistently converging to substantially lower unfaithfulness levels than MAD within a small number of iterations. In contrast, while MAD gradually reduces USR through successive debate rounds, a considerable fraction of unfaithful steps persists even after multiple iterations, indicating that explicitly auditing and repairing localized reasoning failures is more effective than debate-based refinement in preventing the accumulation of unfaithful intermediate reasoning.

\begin{figure*}[t]
\centerline{\includegraphics[width=1.0\textwidth, trim=0 0 0 0,clip]{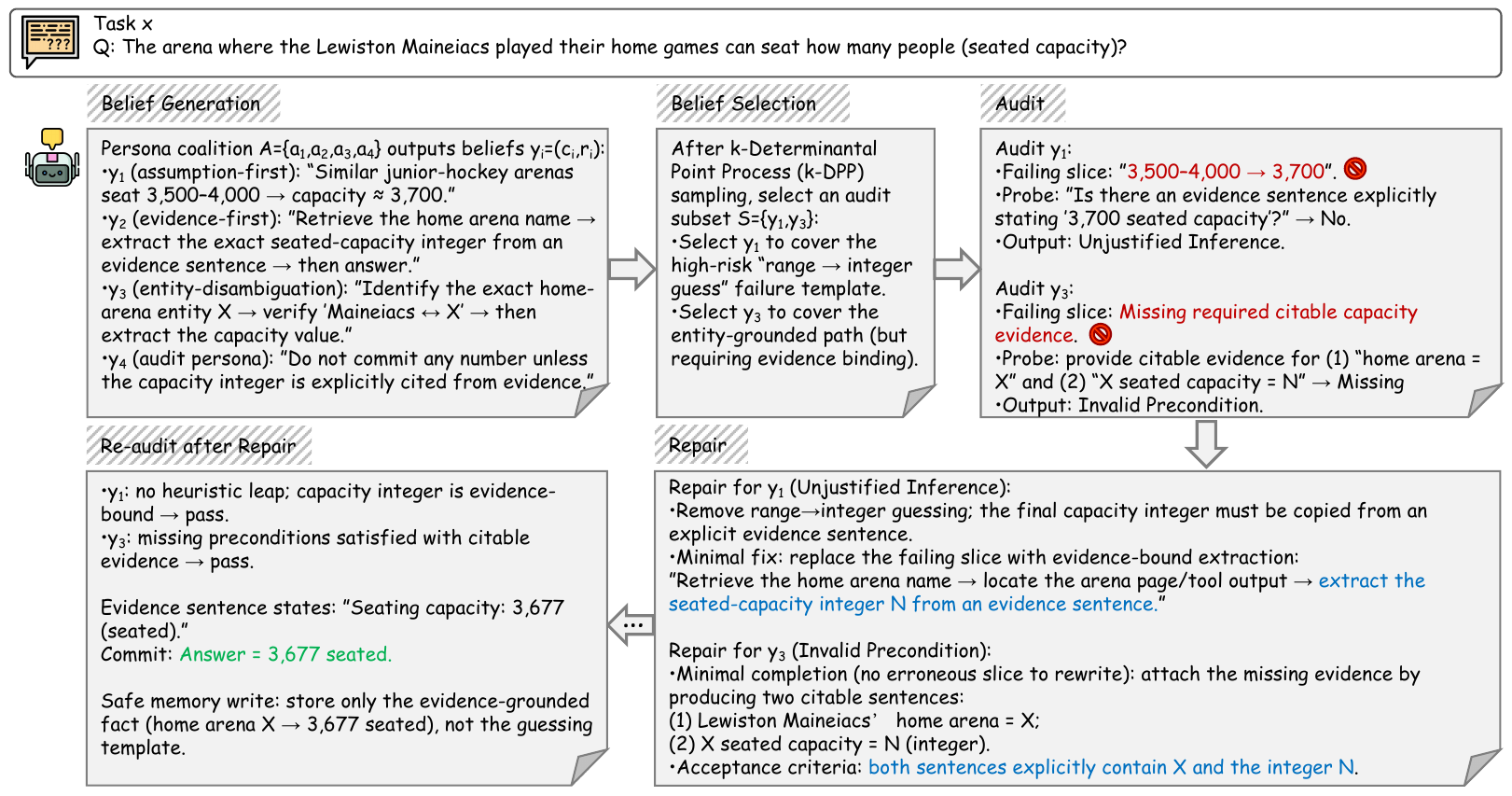}}
\caption{A case study on a multi-hop factual query to correct unjustified inference. The agent initially proposes plausible capacity values based on arena similarity and entity identification, but \textcolor{casered}{without explicit evidence}. \textsc{SAVeR} audits belief candidates, \textcolor{caseblue}{flags unsupported numerical guesses and missing citable capacity statements}, and repairs them by enforcing evidence-bound extraction of the exact seated-capacity integer from retrieved sources. After iterative re-auditing, only the verified capacity value is committed to the \textcolor{casegreen}{final answer} and written to agent memory.}
\label{case_study}
\end{figure*}

\begin{table*}[t]
\centering
\resizebox{1.0\textwidth}{!}{
\begin{tabular}{c|cccccc|cccccc} 
\toprule[1.2pt]
\multirow{2}{*}{\multirowcell{2}{\centering\textbf{Methods}}} & \multicolumn{6}{c|}{\centering\textbf{HotpotQA}} & \multicolumn{6}{c}{\centering\textbf{2WikiMHQA}} \\ \cmidrule[0.5pt](l{1pt}r{0pt}){2-13}

& EM $\uparrow$ & F1 $\uparrow$ & Avg Viol $\downarrow$ & VFR $\uparrow$ & Post-Res $\downarrow$ & USR $\downarrow$ & EM $\uparrow$ & F1 $\uparrow$ & Avg Viol $\downarrow$ & VFR $\uparrow$ & Post-Res $\downarrow$ & USR $\downarrow$ \\ \cmidrule[0.8pt](l{1pt}r{0pt}){1-13}

w/o Persona & 43.2 & 52.4 & 0.49 & 74.55\% & 0.06 & 11.97\% & 47.6 & 55.3 & 0.78 & 65.98\% & 0.10 & 19.24\%  \\ 

w/o k-DPP & 43.3 & 52.2 & 0.64 & 71.47\% & 0.08 & 15.86\% & 47.5 & 55.2 & 0.86 & 61.78\% & 0.14 & 20.52\%  \\ 

w/o Auditing & 43.8 & 52.8 & 1.37 & 42.65\% & -- & 26.74\% & 47.8 & 55.7 & 1.76 & 38.95\% & -- & 29.17\%  \\ 

w/o  Repair & 44.0 & 52.9 & 1.56 & 33.68\% & -- & 37.63\% & 48.1 & 55.6 & 1.83 & 29.17\% & -- & 39.84\%  \\ 

\cellcolor{ablationcolor}\textsc{SAVeR} & \cellcolor{ablationcolor}43.7 & \cellcolor{ablationcolor}52.6 & \cellcolor{ablationcolor}0.37 & \cellcolor{ablationcolor}81.36\% & \cellcolor{ablationcolor}0.05 & \cellcolor{ablationcolor}9.12\% & \cellcolor{ablationcolor}47.7 & \cellcolor{ablationcolor}55.5 & \cellcolor{ablationcolor}0.56 & \cellcolor{ablationcolor}72.34\% & \cellcolor{ablationcolor}0.08 & \cellcolor{ablationcolor}13.84\% \\ 

\bottomrule[1.2pt]
\end{tabular}}
\caption{Ablation Study on the HotpotQA and 2WikiMHQA Benchmarks under LLaMA-3.1-8B.}
\label{ablation}
\vspace{-2pt}
\end{table*}

\subsection{Case Studies and Discussions}
We present a representative case study to illustrate how unfaithful reasoning arises in agentic question answering and how \textsc{SAVeR} mitigates such failures through explicit auditing and repair. As shown in Figure~\ref{case_study}, different personas generate diverse belief candidates, including assumption-driven numerical estimation and evidence-first extraction. During auditing, \textsc{SAVeR} localizes these failures to specific reasoning slices. In our case, one belief commits a numerical estimate (“3,500-4,000 $\rightarrow$ 3,700”) without explicit evidence, which is flagged as an unjustified inference. Another belief identifies the relevant entity correctly but fails to provide a citable sentence linking the arena to its seated capacity, violating required preconditions. Heuristic guessing is replaced with evidence-bound extraction, and missing preconditions are satisfied by explicitly attaching verifiable evidence sentences. The repaired beliefs are then re-audited to ensure all acceptance criteria are met before commitment. As a result, the agent produces a final answer that is fully grounded in cited evidence, preventing the accumulation of unfaithful beliefs in long-horizon reasoning. 

\subsection{Ablation Studies}
In Table~\ref{ablation}, we present the ablation study on HotpotQA and 2WikiMHQA, examining the contribution of each component in \textsc{SAVeR}. Removing any module consistently degrades reasoning faithfulness, while having marginal effects on EM and F1, indicating the effectiveness on the intermediate reasoning quality. Specifically, removing persona generation leads to a noticeable increase in Avg Viol and USR, suggesting the significance of structured reasoning diversity for exposing distinct failure modes. Disabling the $k$-DPP-based belief selection further exacerbates unfaithfulness, highlighting the role of structure-aware diversity in preventing correlated reasoning errors. More severe degradation is observed when auditing or repair is removed: both settings result in substantially higher Avg Viol and USR. These results confirm that audit and constraint-guided repair are essential for effectively reducing unfaithful reasoning.

\section{Conclusion}
In this work, we studied the agent reasoning faithfulness, where coherent reasoning can still violate logical or evidential constraints, and such unfaithful beliefs may propagate and accumulate in agentic systems, leading to systematic behavioral drift. We propose \textsc{SAVeR}, a framework that explicitly verifies intermediate belief states before action commitment. \textsc{SAVeR} generates diverse candidate beliefs, selectively inspects them at the trajectory level, and corrects localized reasoning failures under explicit acceptance criteria, enabling the agent to prevent unsupported inferences from being written to memory. Extensive experiments across multiple benchmarks demonstrate that \textsc{SAVeR} substantially improves reasoning faithfulness while maintaining competitive end-task performance. 


\section*{Limitations}
This work exhibits several limitations worth noting. Firstly, extra computational overhead is introduced by maintaining multiple candidate belief states and performing iterative A-R cycles. Although \textsc{SAVeR} limits audit to a small, structurally diverse subset, the A-R loop remains more expensive than single-pass prompting or lightweight refinement strategies, particularly in tasks with short reasoning chains. Secondly, strict faithfulness enforcement may be unnecessary in simple scenarios and could introduce redundant reasoning operations. While \textsc{SAVeR} is designed to localize and minimally correct unsupported reasoning steps, it currently lacks an explicit mechanism to adapt verification depth to task difficulty. Future work could explore adaptive auditing policies that condition verification on uncertainty or task complexity, enabling agents to dynamically trade off reasoning faithfulness and efficiency.

\section*{Ethical Considerations}
This work aims to improve the faithfulness of internal reasoning in agents by auditing and repairing unsupported intermediate beliefs before they are committed to actions or memory. All experiments are conducted on publicly available datasets, and no additional collection of personal or sensitive data is involved. All models and data are used in accordance with their intended purposes and licenses. Although \textsc{SAVeR} reduces the risk of propagating unfaithful reasoning, it does not guarantee correctness in high-stakes applications such as medical, legal, or safety-critical settings. The auditing and repair procedures rely on the underlying LLM, and biases present in the base model may still affect verification outcomes. Thus, human oversight remains necessary when deploying. 

\section*{GenAI Usage Disclosure}
This work is entirely original and was conducted by the authors. Generative AI tools were not used to produce any content of the work; they were used solely to assist with language refinement and improve clarity and quality of the text.




\bibliography{reference}

\clearpage
\twocolumn

\end{document}